\documentclass[10pt,twocolumn,letterpaper]{article}
\usepackage[pagenumbers]{cvpr}
\usepackage{amsmath,amssymb,booktabs,graphicx,xcolor,longtable}
\definecolor{cvprblue}{rgb}{0.21,0.49,0.74}
\usepackage[breaklinks,colorlinks,allcolors=cvprblue]{hyperref}

\DeclareMathOperator*{\argmin}{arg\,min}
\title{BioCL-POS: Biological Completer-aware Learning with Predictive Observation Selection --- A Wild Ginseng Study}
\author{Yulu Wang\quad Tingting Yao\quad Taolue Liu\\Zhejiang University\\{\tt\small liutaolue@zuaa.zju.edu.cn}}

\hypersetup{pdftitle={BioCL-POS: Biological Completer-aware Learning with Predictive Observation Selection --- A Wild Ginseng Study},pdfauthor={Yulu Wang, Tingting Yao, Taolue Liu}}

\begin{document}
\maketitle
\begin{abstract}
Growers and traders rely on root morphology to assess wild ginseng, yet breakage can obscure important structural traits. We study point cloud completion to provide geometric references for morphological analysis. Does providing more observed points always improve 3D completion? We test whether selected subsets improve recovery of the original missing region while keeping the pretrained completer fixed. We introduce BioCL-POS, an external selector that predicts how the completer responds to candidate inputs. For each candidate, we predict the displacement and distance from shared spatial locations to the nearest points in its completion. We use these spatial responses to estimate recovery quality and supervise them with measurements from actual completions. At inference, we select an input from the observed geometry and run the completer once. We evaluate four frozen completers on controlled gaps in ginseng and tree data at 5\%, 10\%, and 20\% omission, and examine 30\% omission on ginseng. At 5--20\% omission, we outperform full-input completion and all five alternative selectors on both datasets with all four completers. We achieve lower mean reconstruction error and higher mean F-score and coverage. We thus show that selecting observations can improve geometric recovery without retraining the completer, providing more accurate geometric references for subsequent morphological analysis.
\end{abstract}

\section{Introduction}
\label{sec:intro}
Growers, traders, and buyers in China place substantial economic value on wild ginseng. Jilin's government reported RMB 105.29 billion in output across the province's ginseng value chain in 2025, first exceeding RMB 100 billion~\cite{jilin2026ginsengindustry}. A consumer guide published in Jilin Province describes quality grades based on age and morphological traits, including root form and fine-root characteristics~\cite{jilin2024ginseng}. When assessing a damaged specimen, buyers must work with an incomplete record of its original morphology. We aim to recover missing geometric information to support this assessment. Our broader goal is to help reduce economic losses caused by missing information during assessment and trading.

\begin{figure*}[t]
\centering
\includegraphics[width=\textwidth]{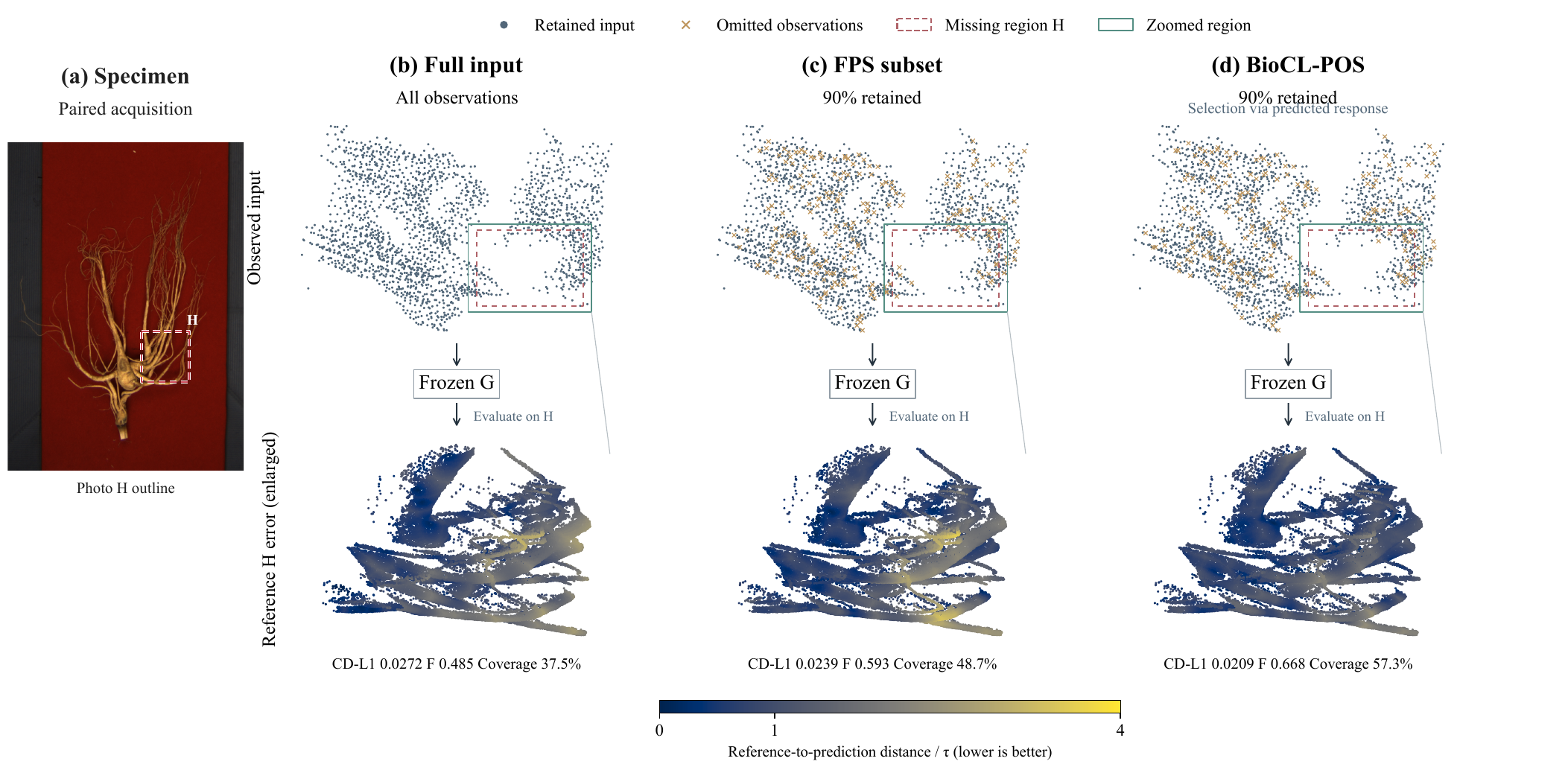}
\caption{Input selection and recovery error on one ginseng test case. In the paired photograph, the dashed red outline marks the location of $H$. We compare Full, farthest point sampling (FPS), and BioCL-POS with the same frozen SnowflakeNet completer $G$ and original missing region $H$. Both subsets retain 90\% of the observations. Gray-blue points enter $G$; orange crosses mark omitted observations. In the point-cloud views, red dashed boxes indicate the projected extent of $H$, and green boxes locate the enlarged reference views below. In the lower row, we color the same measured reference points by their three-dimensional nearest distance to each saved completion, divided by the evaluation threshold $\tau$. Blue indicates smaller error and yellow larger error, with a shared color scale. The lower shapes are measured references, not generated completions. F denotes F-score. We compute all metrics over the complete missing region and all predicted points.}
\label{fig:application}
\end{figure*}

We study point cloud completion to infer missing root geometry from the remaining observations. We examine how input selection changes recovery of the missing region (Fig.~\ref{fig:application}). We could also use the completed geometry to extract morphological features, compare specimens, and support subsequent automated grading. To evaluate geometric recovery, we withhold local regions from measured scans and compare predictions with the withheld measurements.

Researchers learn local geometric relationships through dynamic graphs and point attention~\cite{dgcnn,pointtransformer}, and use increasingly expressive models for point cloud completion. With CRA-PCN and AdaPoinTr, they combine geometric information at multiple resolutions and adapt the generation of missing points to the input~\cite{cra,ada}. With SeedFormer and SnowflakeNet, they carry local geometric information into progressively finer predictions~\cite{seed,snow}. To recover a missing region, we rely on a trained completer to infer geometry from the surrounding observations. We might therefore expect to obtain the best recovery by supplying every available observation. But does a fixed completer always recover the gap best from the full input? We use the complex, branching morphology of wild ginseng to test this question, comparing full and selected inputs while keeping the model and original gap unchanged. We examine whether any subset improves recovery, then ask how to identify such a subset from the observed geometry.

To identify useful candidates, we measure geometric error with the symmetric Chamfer distance~\cite{psgn}, using the unsquared Euclidean form~\cite{pcn}, which we denote CD-L1. Lower values indicate closer agreement with the reference. We also measure precision and coverage as the fractions of predicted and reference points, respectively, that lie within a distance threshold of the other point set. We report their harmonic mean as F-score. In the saved sets of candidate inputs for ginseng, each subset omits 10\% of the unique observed coordinates. Across the four completers, we find a subset that reduces CD-L1 while preserving F-score and coverage within the stated tolerance in 66.9--81.2\% of the recorded evaluation settings. We compute these percentages by giving each recorded group equal weight (Sec.~\ref{sec:mechanism}).

To exploit such an opportunity, we must choose the input before seeing its completion. Researchers have already learned to sample points for downstream tasks. The authors of SampleNet and LTA-PCS motivate sampling through computational or storage costs, while also seeking to preserve task performance~\cite{samplenet,lta}. With SAMBLE, the authors address the balance between local detail and global uniformity to improve sampling for downstream tasks~\cite{samble}. We focus on predicting how a particular completer responds to each candidate input. We supervise predictions of this spatial response and use them to estimate recovery quality.

We introduce BioCL-POS, a completer-aware selector trained for a specific pretrained completer whose parameters we keep fixed (Fig.~\ref{fig:framework}). We construct equal-sized candidates by switching which observed point we retain within local groups of point pairs. We use the same switches to select the group features for predicting each candidate's completion. At shared spatial locations, we predict the displacement and distance to its nearest output points. We can thus compare unordered completions without matching their output points. Within the response predictor, we estimate recovery quality from these spatial responses and rank candidates. During training, we supervise response predictions with descriptors extracted from actual completions and quality estimates with the withheld measurements. At inference, we supply the selected measured coordinates to the completer for one final completion. We retain all original measurements in the digital record.

We evaluate primarily on wild ginseng scans and use tree data to examine another form of plant branching structure. We train selectors for four frozen completers. We compare against completion from all observed points (Full) and the average results from four randomly sampled subsets (RandomExpected). We also compare with farthest point sampling (FPS), which repeatedly selects the point farthest from those already selected~\cite{pointnetpp}, and the three learned sampling methods above. We adapt these samplers to the same completion task. We assess both how often useful candidates exist and how reliably we select them. We compare the same network trained with and without supervision of its predicted responses, keeping the training budget unchanged. We also permute candidate scores to examine how their association with input choices affects selection and gap recovery. We measure the cost of the complete pipeline and examine performance across scan sides and missing-region conditions.

We make three contributions:
\begin{itemize}
\item We investigate whether more observed points lead to better 3D completion, using wild ginseng as a test case and keeping the original recovery target fixed.
\item We develop BioCL-POS to connect local input choices with a frozen completer's spatial response at shared locations and select inputs by predicted recovery quality.
\item We construct a wild ginseng completion benchmark from our own scans, with controlled local gaps, defined evaluation partitions, and a common protocol for comparing input selectors and frozen completers. We will publicly release the dataset upon acceptance.
\end{itemize}

\section{Related Work}
\label{sec:related}
\paragraph{Point Cloud Completion.}
Researchers develop completion models by choosing how to represent missing geometry and carry observed structure into it. They use folding-based decoders for point-cloud reconstruction~\cite{foldingnet} and generate surface patches for completion~\cite{pcn,msn,sparenet}. They also organize point generation hierarchically, as in TopNet and SnowflakeNet~\cite{topnet,snow}. Researchers make spatial structure explicit through intermediate grids and successive point displacements~\cite{grnet,pmpnet}. During refinement, they preserve local information through skip-attention, patch seeds in SeedFormer, and multiresolution features in CRA-PCN~\cite{sanet,seed,cra}. With attention, they relate observed and missing geometry through point-group features and point relationships~\cite{pointr,proxyformer,pointattn}. In AdaPoinTr, the authors further adapt decoder queries, the features used to generate missing points, to the input~\cite{ada}. Researchers also exploit projected depth views and geometric self-similarity~\cite{svdformer,pointsea}, model uncertainty in plausible completions~\cite{vrcnet,shapeformer}, and learn from single partial observations through self-supervision~\cite{p2c}. For application-specific reconstruction, researchers predict planar primitives for polygonal surfaces~\cite{paco} and inpaint projected image patches to restore archaeological point clouds~\cite{archaeology}. We focus on selecting measured inputs for a fixed completer to recover the same original gap.

\paragraph{Task-Oriented Point Cloud Sampling.}
Researchers select points to preserve spatial coverage with FPS~\cite{pointnetpp}, salient outlines with APES, or relations within and between point clusters with attention discriminant sampling~\cite{apes,ads}. With SAMBLE and LTA-PCS, they respectively balance local detail with global uniformity and combine geometric objectives with semantic features across tasks~\cite{samble,lta}. For a known downstream task, they can instead learn sampling through its loss: S-NET generates points and matches them to observations, while SampleNet passes task gradients through soft projection~\cite{snet,samplenet}. APSNet extends task-oriented sampling through sequential decisions conditioned on earlier selections~\cite{apsnet}. Researchers also estimate point-dropping sensitivity through recognition loss~\cite{saliency}, or adjust sampled coordinates for robustness to noisy observations in PointASNL~\cite{pointasnl}. We retain original measured coordinates and compare equal-sized subsets by their predicted completion responses, using recovery of the missing region as the selection objective.

Researchers compare choices before execution by predicting future measurements~\cite{dfp}, action success~\cite{where2act}, or resulting shapes~\cite{sculptbot}. In next-best-view planning, they use such predictions to choose sensor positions for reconstruction. They estimate candidate-view gains directly in PC-NBV~\cite{pcnbv}, or derive coverage from predicted occupancy and visibility in SCONE~\cite{scone}. They also learn occupancy and view selection online from color images in MACARONS~\cite{macarons}, and guide view selection with predicted complete geometry in Pred-NBV~\cite{prednbv}. We apply prediction-guided selection to observations already acquired. By describing a fixed completer's response at shared spatial locations, we learn to compare candidate inputs before the final completion call.

\section{Problem Formulation}
\label{sec:problem}
We study observation selection for recovering a missing region with a fixed point cloud completer. Let $X\subset\mathbb{R}^3$ denote the available observations and $H\subset\mathbb{R}^3$ the reference points in the original missing region, with $X\cap H=\varnothing$. For a selected subset of $X$, we use a pretrained completer $G_\theta$ to predict the missing geometry. We keep its parameters $\theta$ fixed and evaluate every input choice against the same target $H$. During training and evaluation, we use $H$ for supervision and assessment; at deployment, we select inputs from the available observations.

For a prescribed omission ratio $\rho$, we construct a set of candidate actions $\mathcal A_\rho(X)$ from the observed geometry. For each action $a$, we omit a set of measured points $M_a\subset X$ from the input. We then supply the retained subset $S_a$ to the completer and obtain the predicted points $P_a$:
\begin{equation}
S_a=X\setminus M_a,\qquad P_a=G_\theta(S_a).
\label{eq:task}
\end{equation}
At each $\rho$, we compare candidates with the same number of retained points.

Within this candidate set, we seek the input that minimizes recovery error. We denote CD-L1 by $D(P,H)$ and compute it by averaging the nearest-point distances in the two directions, with equal weight for each direction. We define the ideal choice as
\begin{equation}
a^\dagger=\argmin_{a\in\mathcal A_\rho(X)}D\!\left(G_\theta(S_a),H\right).
\label{eq:ideal_selection}
\end{equation}
We also evaluate F-score and coverage to assess completion quality. To select an input before observing its completion, we learn to estimate the recovery error for each candidate. In the following section, we describe how we predict the completer's geometric response and use it to guide this estimate.

\section{BioCL-POS}
\label{sec:method}
\begin{figure*}[t]
\centering
\includegraphics[width=\textwidth]{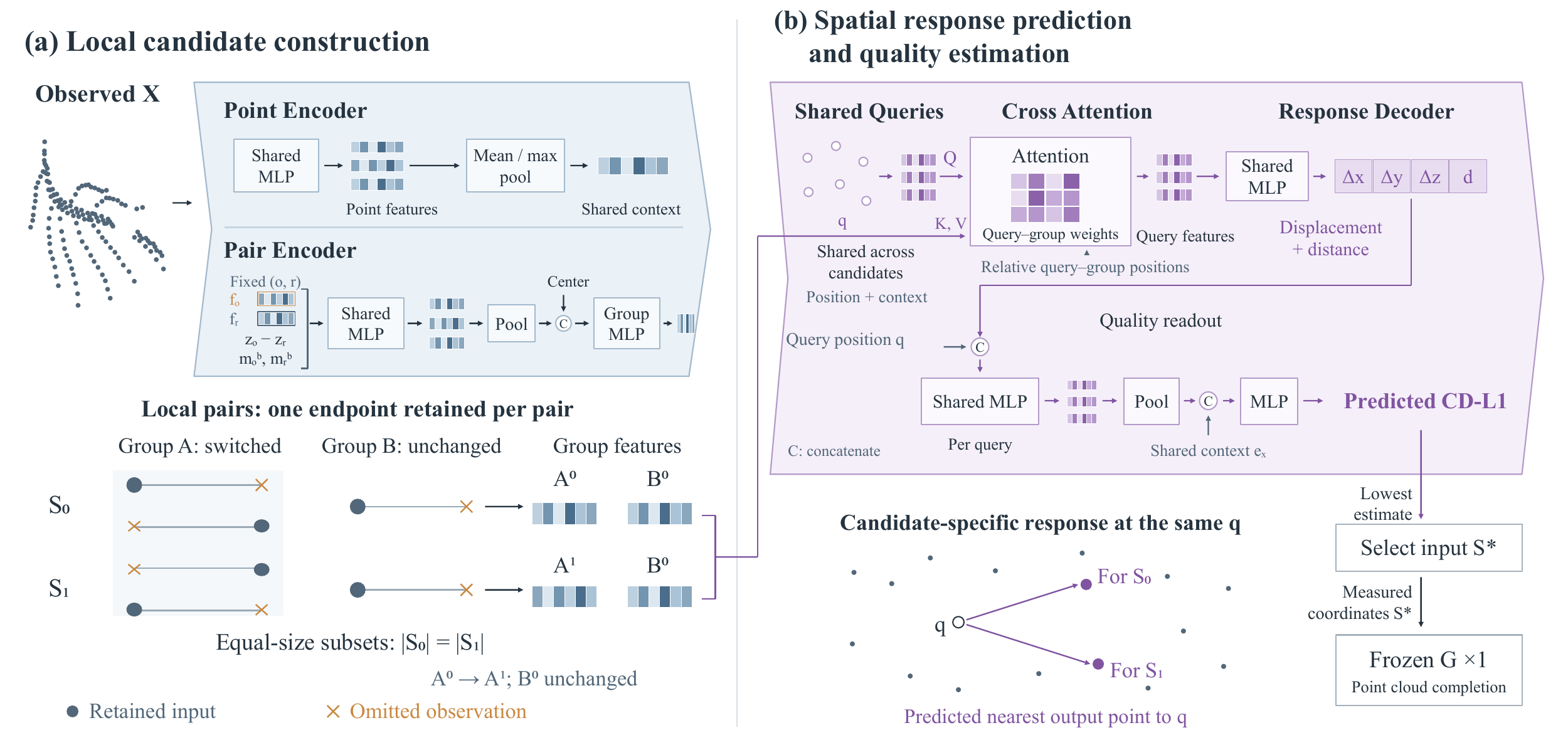}
\caption{Overview of BioCL-POS. (a) We construct equal-sized candidate inputs by switching the retained point in each pair within a group, and encode these choices as group features. In the example, we switch both pairs in group A and keep group B unchanged. (b) Within one module, we predict the nearest points in the frozen completer's output at shared query locations and estimate completion quality through a readout of these responses. We select the candidate with the lowest predicted CD-L1 and pass its original measured coordinates to $G$ for one final completion. We illustrate predicted spatial responses in purple. We supervise response predictions with offline completions and quality estimates with errors measured against $H$.}
\label{fig:framework}
\end{figure*}

We organize BioCL-POS into two parts: candidate construction and encoding, and spatial response prediction with quality estimation (Fig.~\ref{fig:framework}). We construct candidates by switching retained endpoints within groups of nearby point pairs. We use the same group states to assemble features for predicting the completer's spatial responses at shared locations. Within this second part, we estimate recovery quality from the predicted responses. We then choose the candidate with the lowest predicted CD-L1 and complete its measured coordinates in one call to the fixed completer. We learn the responses from offline completions and the quality estimates from their errors against the original missing region.

\subsection{Candidate construction and encoding}
We construct candidate inputs through local replacements, retaining a nearby observation for each omitted point. To do so, we form disjoint pairs of nearby observations, with each point belonging to at most one pair. We count the unique observed coordinates as $n=|X|$ and set the omission budget to $k_\rho=\max(1,\operatorname{round}(\rho n))$ for $\rho>0$. To construct $k_\rho$ pairs, we process connections between neighboring points from shortest to longest and accept a pair when neither point has been assigned. We then vary which endpoint we retain from each pair and keep all unpaired observations. We thus compare local input choices while keeping the number of retained points fixed at $n-k_\rho$.

We group nearby pairs so that one binary choice can specify several local replacements within the same spatial region. We apply FPS to the pair midpoints to select up to $N_g$ group centers, then assign each pair to the nearest center. For each endpoint, we measure its distance to its second nearest neighbor in $X$. We initially omit the endpoint with the smaller distance. We encode each group's state with one bit $b_{ag}$: 0 keeps the initial choices, and 1 exchanges the retained and omitted endpoints in every pair. We construct $S_a$ by collecting the retained points from all groups together with the unpaired observations. We evaluate $N_c$ predefined masks, set bits for empty groups to zero, and retain the first occurrence of each distinct mask. We denote these nonduplicate candidates by $\mathcal A_{\rho,\mathrm{valid}}(X)$. In Fig.~\ref{fig:framework}, for example, we obtain $S_1$ from $S_0$ by switching both pairs in group A and leaving group B unchanged.

We use the same mask to construct each candidate and assemble its features for response prediction. Within each pair, we encode the retained and omitted roles so that the representation specifies the local choice. We first normalize the observed coordinates as $z_i=(x_i-c_X)/s_X$, where $c_X$ is the observation centroid and $s_X$ is its bounding-box diagonal. With the Point Encoder, we apply a shared multilayer perceptron (MLP) to each $z_i$ to obtain a point feature $f_i$, following the pointwise encoding used in PointNet~\cite{pointnet}. We concatenate the mean and maximum point features into $e_X$, which summarizes the full observation. For each pair $\ell$, we fix the endpoint order $(o,r)$ by the initial omitted and retained roles. We append retention flags $m^b=(b,1-b)$ to encode each subset state with the Pair Encoder:
\begin{equation}
u_\ell^{\,b}=\operatorname{MLP}_{\mathrm{pair}}\!\left([f_o,f_r,z_o-z_r,m_o^b,m_r^b]\right).
\label{eq:pairfeature}
\end{equation}
We then pool the pair features within each group $g$, concatenate its normalized center $c_g$, and apply another MLP to obtain the group feature $t_g^{\,b}$. We compute the group states once and reuse them across candidates. For candidate $S_a$, we use the same mask that constructs its input to select $t_g^{\,b_{ag}}$ from each group. We pass these features and their spatial centers to the response predictor.

\subsection{Spatial response prediction and quality estimation}
To compare how the completer responds to different group states, we describe its output at shared spatial locations. We can then compare unordered completions through the same query locations without matching output points across candidates. At each location, we describe the completion by the displacement to its nearest output point and the corresponding Euclidean distance. Let $q_j$ denote a query location and $\widetilde P_a$ the completion of candidate $S_a$, expressed in the normalized coordinate frame. We define the nearest output point and the spatial response at $q_j$ as
\begin{align}
p^*_{aj}&=\argmin_{p\in\widetilde P_a}\|p-q_j\|_2,\\
r_{aj}&=\big[p^*_{aj}-q_j,\ \|p^*_{aj}-q_j\|_2\big].
\label{eq:response}
\end{align}
We thus describe each response with three displacement components and one distance.

We construct the query locations from the normalized observation. For an even query count $J$, we select $J/2$ observed points using FPS and add the midpoint between each selected point and the observation centroid, obtaining $J$ locations. We construct this set once from $X$, giving all candidates the same spatial reference. For candidate $S_a$, we collect the responses at these locations as $R_a=(r_{aj})_{j=1}^{J}$.

At each query location, we use cross-attention~\cite{attention} to read the candidate's selected group features. We encode each query location together with $e_X$ as $e_j=\phi_q([q_j,e_X])$, where $\phi_q$ is a learned embedding function. We use $e_j$ as the query feature and provide the selected group features $T_a=\{t_g^{\,b_{ag}}\}_g$ as keys and values. We also incorporate $q_j-c_g$ into the attention weights, relating each group choice to the location where we predict its response. We obtain $h_{aj}$ through cross-attention. We then use a shared MLP in the Response Decoder to predict the displacement and distance:
\begin{align}
h_{aj}&=\operatorname{CrossAttn}\!\left(e_j,T_a;\{q_j-c_g\}_g\right),\\
\widehat r_{aj}&=\operatorname{MLP}_{r}(h_{aj}).
\label{eq:predict}
\end{align}
We share the embedding function, attention layer, and decoder across all candidates and query locations.

Within the response prediction module, we estimate recovery quality from the predicted spatial responses. During training, we learn this mapping from errors measured against the reference region $H$. We encode each response together with its query coordinates using a shared MLP $\phi_v$. We average these features over the query locations and concatenate the result with $e_X$. Through a quality readout, we map this representation to three predicted metrics:
\begin{equation}
\widehat v_a=\sigma_v\!\left(\operatorname{MLP}_{v}\!\left(
\left[\frac{1}{J}\sum_{j=1}^{J}\phi_v([q_j,\widehat r_{aj}]),e_X\right]\right)\right).
\label{eq:value}
\end{equation}
We denote the outputs by $\widehat v_a=(\widehat d_a,\widehat c_a,\widehat f_a)$, corresponding to normalized CD-L1, coverage, and F-score. Through $\sigma_v$, we apply softplus to keep the predicted distance positive and sigmoid to constrain predicted coverage and F-score to $(0,1)$. We share this readout, the query locations, and the observation context across candidates. Candidate differences reach the quality readout through the predicted responses.

For subset-only selection in the matched scoring and mechanism analyses, we minimize predicted CD-L1 over valid subsets:
\begin{equation}
a^*=\argmin_{a\in\mathcal A_{\rho,\mathrm{valid}}(X)}\widehat d_a,
\qquad \widehat H=G_\theta(S_{a^*}).
\label{eq:select}
\end{equation}
Using the selected mask, we retrieve the retained points from $X$ in their original measured coordinates and pass them to $G_\theta$ for one completion. Test comparisons additionally include Full and apply the quality screen in Sec.~\ref{sec:experimental_setup}.

\subsection{Learning from frozen-model responses}
We obtain training supervision from the frozen completer's outputs and the measured reference region. For each training case, we construct candidate subsets from $X$ and run the pretrained completer offline to obtain $P_a=G_\theta(S_a)$. We express each output in the normalized coordinate frame and extract its responses $R_a$ at the shared query locations using the definition above. We also compare $P_a$ with the reference points $H$ to compute CD-L1, coverage, and F-score. We divide the measured CD-L1 by $s_X$ and form the quality target as
\begin{equation}
v_a=(\widetilde d_a,c_a,f_a),\qquad \widetilde d_a=\frac{D(P_a,H)}{s_X},
\label{eq:quality_target}
\end{equation}
where $c_a$ and $f_a$ denote coverage and F-score, respectively. We store these response and quality targets with their candidate inputs before training. We then optimize the selector against these targets while keeping the completer's parameters fixed.

We train the selector to predict spatial responses, estimate completion quality, and order candidate inputs through three losses:
\begin{equation}
\mathcal L=\mathcal L_{\mathrm{response}}+\mathcal L_{\mathrm{value}}+\mathcal L_{\mathrm{rank}}.
\label{eq:loss}
\end{equation}
For $\mathcal L_{\mathrm{response}}$, we fit each predicted response $\widehat r_{aj}$ to its target $r_{aj}$ using smooth-$L_1$ loss, which penalizes small errors quadratically and larger errors linearly. We average this loss over valid candidates, query locations, and response components. For $\mathcal L_{\mathrm{value}}$, we apply the same loss to $\widehat v_a$ and $v_a$, averaging over valid candidates and the three metrics. To supervise the ordering used for selection, we compare candidates from the same case with a pairwise logistic ranking loss~\cite{ranknet}. Let $\mathcal E$ contain valid ordered pairs $(a,b)$ whose target errors differ by more than a tolerance $\epsilon_{\mathrm{rank}}$. We define
\begin{equation}
\mathcal L_{\mathrm{rank}}=\frac1{|\mathcal E|}\sum_{(a,b)\in\mathcal E}
\log\left(1+e^{-\operatorname{sgn}(\widetilde d_a-\widetilde d_b)(\widehat d_a-\widehat d_b)}\right),
\label{eq:rank}
\end{equation}
where $\operatorname{sgn}$ denotes the sign of the target difference. Through this term, we train the selector to assign a lower predicted distance to the candidate with lower recovery error. We set the ranking loss to zero when no eligible pairs exist. We train on cases with nonempty reference regions.

\section{Experiments}
\label{sec:experiments}
We report six experiment groups: recovery gains, ablations, selection mechanisms, inference cost, stability, and geometric reliability.

\subsection{Experimental Setup}
\label{sec:experimental_setup}
\paragraph{Data and comparisons.}
We split the 728 wild ginseng specimens into 50\% training, 20\% validation, and 30\% test data, keeping each specimen's views and gap cases together. We tune on validation data and report main comparisons on the test split. We also use a complementary 160-model subset of Synthetic-Trees-II spanning ten tree species~\cite{synthetic_trees_ii}. We withhold a fixed local region $H$ per scan across input choices. For trees, we use 96 models for selector fitting, 24 for internal evaluation, and 40 for testing. We freeze CRA-PCN, AdaPoinTr, SeedFormer, and SnowflakeNet~\cite{cra,ada,seed,snow}, each predicting 2,048 points for $H$. We compare BioCL-POS with Full, RandomExpected (four random subsets), FPS, SampleNet~\cite{samplenet}, LTA-PCS~\cite{lta}, and SAMBLE~\cite{samble}, sharing completer checkpoints and adapters. We test 5\%, 10\%, and 20\% omission on both datasets and 30\% on ginseng. We examine four training seeds and four completer input orders.

\paragraph{Configuration and decision rules.}
We use up to $N_g=12$ groups, $N_c=16$ candidate masks, and $J=64$ query locations (32 observed points and 32 midpoints). We use one cross-attention layer, set $\epsilon_{\mathrm{rank}}=10^{-8}$, and weight the three losses equally. The final input is selected from the candidate subsets and the full observation in test comparisons. We exclude candidates whose predicted F-score or coverage is more than 0.001 below the corresponding prediction for Full, then minimize predicted CD-L1 over the remaining options. The omission budget applies to candidate subsets. The matched scoring and mechanism analyses instead select among the same 16 subsets by predicted CD-L1, as in Eq.~\ref{eq:select}.

\paragraph{Metrics and statistics.}
\label{sec:metrics}
We report CD-L1, F-score, and coverage. We measure unsquared distances in stored coordinates; proximity thresholds are $0.02\operatorname{diag}(X)$ for ginseng and $0.02$ for trees. For each run, we average cases within each group and view, then average views within groups. We average runs within groups and weight groups equally. We pair methods by case, point order, and completer checkpoint, resampling whole groups for 95\% confidence intervals. Relative CD gains average within-group reductions; relative F-score and coverage gains compare weighted means. A \emph{protected improvement} reduces CD-L1 against Full without lowering F-score or coverage by more than 0.001. We call a candidate that meets this criterion an improving candidate.

\subsection{Completion with Selected Inputs}
\label{sec:main_effect}
\begin{table*}[t]\centering
\caption{Recovery on the test set with a 10\% candidate omission budget and Full included among the input choices. Absolute CD-L1 means for Full and BioCL-POS come directly from the same current group summaries (stored coordinate units). Gain columns report relative CD-L1 reduction or F-score/coverage increase (\%), with 95\% paired intervals. Relative CD gains average within-group ratios, so they need not equal the ratio of the two absolute means. The last column compares with the strongest CD reference; the reference pool includes Full, DIRECT-0, and five selectors.}
\label{tab:main}
\small\setlength{\tabcolsep}{4pt}\renewcommand{\arraystretch}{1.14}
\begin{tabular}{llrrrrrr}
\toprule
Dataset & Completer & Full CD & Ours CD & CD gain & F gain & C gain & CD vs strongest \\ \midrule
Ginseng & CRA-PCN & 0.04966 & 0.03841 & \shortstack{23.01\\{\scriptsize [20.63,25.21]}} & \shortstack{32.96\\{\scriptsize [26.74,39.85]}} & \shortstack{15.49\\{\scriptsize [13.67,17.35]}} & \shortstack{21.05\\{\scriptsize [18.52,23.38]}} \\
Ginseng & AdaPoinTr & 0.09845 & 0.08986 & \shortstack{8.64\\{\scriptsize [7.91,9.33]}} & \shortstack{97.57\\{\scriptsize [80.83,119.78]}} & \shortstack{13.64\\{\scriptsize [10.35,17.41]}} & \shortstack{8.48\\{\scriptsize [7.97,9.00]}} \\
Ginseng & SeedFormer & 0.05712 & 0.04795 & \shortstack{15.73\\{\scriptsize [14.47,17.29]}} & \shortstack{37.24\\{\scriptsize [30.16,45.69]}} & \shortstack{14.82\\{\scriptsize [12.31,18.01]}} & \shortstack{13.76\\{\scriptsize [12.69,14.89]}} \\
Ginseng & SnowflakeNet & 0.05720 & 0.04804 & \shortstack{16.19\\{\scriptsize [14.58,17.94]}} & \shortstack{37.60\\{\scriptsize [29.77,46.74]}} & \shortstack{12.54\\{\scriptsize [10.27,15.14]}} & \shortstack{16.19\\{\scriptsize [14.58,17.94]}} \\
Trees & CRA-PCN & 0.04733 & 0.04145 & \shortstack{11.83\\{\scriptsize [10.52,13.17]}} & \shortstack{26.79\\{\scriptsize [17.30,42.38]}} & \shortstack{10.06\\{\scriptsize [6.91,14.56]}} & \shortstack{11.15\\{\scriptsize [10.48,11.90]}} \\
Trees & AdaPoinTr & 0.04241 & 0.03790 & \shortstack{10.29\\{\scriptsize [9.30,11.31]}} & \shortstack{20.64\\{\scriptsize [14.81,29.39]}} & \shortstack{16.76\\{\scriptsize [12.74,22.14]}} & \shortstack{10.46\\{\scriptsize [9.60,11.45]}} \\
Trees & SeedFormer & 0.04858 & 0.04507 & \shortstack{8.23\\{\scriptsize [6.58,10.55]}} & \shortstack{21.08\\{\scriptsize [13.43,34.26]}} & \shortstack{7.72\\{\scriptsize [5.41,10.76]}} & \shortstack{7.88\\{\scriptsize [6.70,9.79]}} \\
Trees & SnowflakeNet & 0.04713 & 0.04286 & \shortstack{9.42\\{\scriptsize [8.10,11.11]}} & \shortstack{29.49\\{\scriptsize [18.63,49.69]}} & \shortstack{11.93\\{\scriptsize [8.30,16.96]}} & \shortstack{9.63\\{\scriptsize [8.54,10.98]}} \\
\bottomrule\end{tabular}
\end{table*}

We improve average recovery through input selection while keeping the completer and original gap fixed (Table~\ref{tab:main}). With 10\% candidate omission, we reduce CD-L1 by 8.64--23.01\% on ginseng and 8.23--11.83\% on trees, with higher mean F-score and coverage for every completer. For CRA-PCN on ginseng, mean CD-L1 decreases from 0.04966 to 0.03841, while F-score and coverage increase by 32.96\% and 15.49\%. We also improve all three mean metrics over each of the five alternative selectors at the same subset budget. We assess scoring under a common rule below.

In Fig.~\ref{fig:application}, both FPS and BioCL-POS retain 90\% of the observations. Coverage increases from 0.3750 with Full to 0.4866 with FPS and 0.5725 with BioCL-POS. The lower panels locate these changes through error on the same reference points, not through rendered predictions. We bring more of the measured gap within the distance threshold, illustrating why input count alone does not determine recovery quality.

\begin{figure*}[t]
\centering
\includegraphics[width=\textwidth]{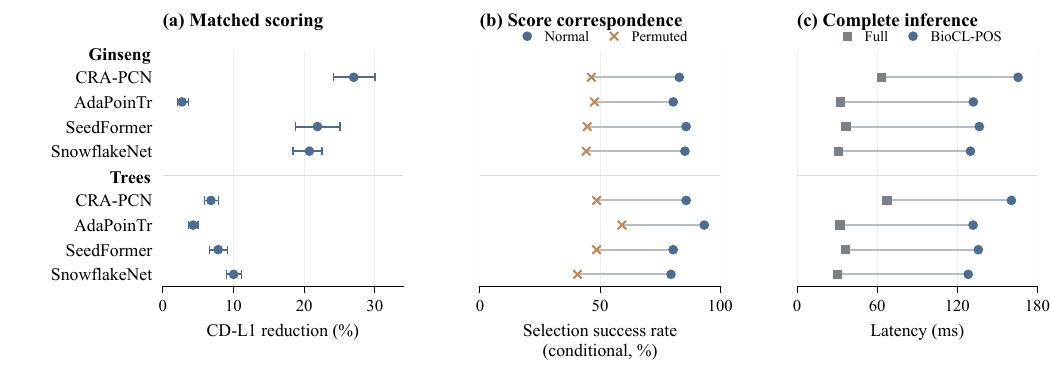}
\caption{Scoring, selection, and cost at 10\% candidate omission. (a) CD-L1 reduction over control scores on the same 16 internal candidates under the same minimum-CD rule; bars show 95\% paired intervals. (b) Selection success rates before and after score permutation, conditional on an improving candidate being available. (c) Complete inference medians on four fixed internal cases per setting, with five timed repeats; Full provides the completion-only reference.}
\label{fig:evidence}
\end{figure*}

\subsection{Ablation and Matched Scoring}
\label{sec:ablation}
We compare our scores with control scores on the same 16 internal candidates, selecting the subset with minimum predicted CD-L1. With cases, completers, seeds, and point orders paired, we reduce CD-L1 by 2.73--27.02\% on ginseng and 4.29--10.03\% on trees, while improving mean F-score and coverage for every completer (Fig.~\ref{fig:evidence}a). Better scoring therefore improves selection within the same available geometry.

For \textit{Without response supervision} (\texttt{DIRECT\_0}), we set the response-fitting loss to zero while retaining the integrated response predictor and quality readout. We train this control through the quality and ranking losses, keeping other training settings unchanged. We outperform it on all three metrics in the test comparison, where BioCL-POS uses the test decision rule. We evaluate scoring under the shared internal rule above and report the test comparison separately.

\subsection{From Candidate Opportunities to Selected Gains}
\label{sec:mechanism}
We first measure the fraction of conditions with at least one improving candidate. In the internal 16-subset banks, this rate is 66.91--81.25\% on ginseng and 58.85--73.44\% on trees. Among conditions with an improving candidate, we successfully select one in 80.46--85.79\% and 79.51--93.36\%, respectively. We report this conditional proportion as the \emph{selection success rate}. We define \emph{selection regret} as the chosen CD-L1 minus the minimum measured candidate CD-L1. Relative to control scores, we reduce regret by 90.97--93.92\% on ginseng and 93.35--95.31\% on trees.

Beyond the selected candidate, we assess the full ordering against measured candidate errors. On ginseng, Spearman rank correlations range from 0.845 to 0.896.

We then permute scores among candidates, preserving score values and completion outcomes. On ginseng, this lowers the selection success rate by 32.80--41.14 percentage points (Fig.~\ref{fig:evidence}b). Across both datasets and every completer, CD-L1 and regret increase while F-score and coverage decrease. We thus connect the benefit to the score--candidate correspondence. These score-level tests do not intervene on network responses.

\subsection{Complete Inference Cost}
\label{sec:efficiency}
On an RTX 5090, the complete pipeline takes 129.85--165.53\,ms on ginseng, including 99.72--100.59\,ms for selection and dispatch; Full takes 30.85--63.31\,ms (Fig.~\ref{fig:evidence}c). One final completer call therefore still incurs additional selection cost. Across both datasets, we reduce latency by 20.29--29.22\% relative to FPS, SampleNet, and LTA-PCS, but spend 4.32--9.68\% more time than SAMBLE alongside higher test-set recovery quality. Timing uses fixed internal cases; quality comparisons use test records.

\subsection{Stability and Geometric Reliability}
\label{sec:generalization}
\label{sec:reliability}
We improve all three mean metrics over Full at 5\%, 10\%, and 20\% omission for every completer on both datasets. We obtain the largest ginseng CD-L1 reductions at 20\% omission for all four completers. With AdaPoinTr, for example, the gain increases from 6.69\% at 5\% omission to 10.83\% at 20\%. On trees, SeedFormer favors 5\% omission, while the other completers favor 10\%. We therefore use the budget comparison to identify the useful extent of selection for each setting. At 30\% omission on ginseng, all four completers have higher mean relative CD-L1 error than Full. The best omission level depends on the dataset and completer.

At 5\%, 10\%, and 20\% omission, every training seed and point order has a positive marginal mean CD-L1 gain after averaging over the other axis.

At 10\% candidate omission, we achieve protected improvements in 59.77--85.54\% of weighted ginseng conditions and 68.44--82.03\% of tree conditions. We define non-degradation as CD-L1 no higher than Full, with neither F-score nor coverage decreasing by more than 0.001. We obtain non-degradation rates of 94.86--98.52\% on ginseng and 97.50--99.06\% on trees. Selecting Full counts toward non-degradation, not improvement; with CRA-PCN on ginseng, we select Full in 31.13\% of conditions. Severe losses---CD-L1 increasing by more than 10\% or F-score/coverage decreasing by more than five percentage points---occur in fewer than 0.5\% of weighted conditions for every dataset--completer pair.

\section{Discussion and Limitations}
\label{sec:discussion}
We interpret the gains over Full as evidence for adapting the observed context to a fixed completer. We assess this adaptation through recovery of the original gap, so the usefulness of a subset depends on the model and target it serves.

For ginseng, we can connect geometric recovery to morphological analysis through measurements extracted from a completed scan. We would preserve the original observations and distinguish the inferred points when computing these measurements. Using our controlled-gap scans, we could compare measurements before and after completion against those obtained from the original scans. We could then identify morphological features whose estimates improve after completion and examine their use in specimen comparison and automated grading.

We search within a predefined set of local group switches and describe each candidate's completion at 64 shared spatial locations. We therefore assess input choices through the alternatives and spatial responses represented by this design. We could extend this approach by adapting candidate construction and query placement to each observation, concentrating computation where alternative inputs produce different responses. We also train the predictor separately for each frozen completer. Learning to adapt an existing predictor to another completer offers a further direction for reducing repeated training.

\section{Conclusion}
We introduce BioCL-POS to select observed points by predicting how a frozen completer responds to candidate inputs. We use these predicted spatial responses to estimate recovery quality and choose an input before one final completion call. We also construct a controlled-gap benchmark from measured wild ginseng scans, allowing us to compare input choices against the same withheld geometry. Across four frozen completers, we find that selected subsets can improve recovery over Full on wild ginseng, with supporting results on tree structures.

\clearpage
\section*{LLM Usage Statement}
We used OpenAI (ChatGPT / GPT) for assistance with writing and editing, related work, figures, and reviewer-style feedback. We reviewed, edited, or approved tool-generated code and experimental analyses. The authors take full responsibility for all submitted content, regardless of any LLM or AI-tool use.

{\small
\bibliographystyle{ieeenat_fullname}
\bibliography{references}
}
\end{document}